\documentclass{article}

\PassOptionsToPackage{numbers, compress}{natbib}
    \usepackage[preprint]{neurips_2020}

\usepackage[utf8]{inputenc} % allow utf-8 input
\usepackage[T1]{fontenc}    % use 8-bit T1 fonts
\usepackage{url}            % simple URL typesetting
\usepackage{booktabs}       % professional-quality tables
\usepackage{amsfonts}       % blackboard math symbols
\usepackage{amsmath}
\usepackage{makecell}
\usepackage{graphicx} 
\usepackage{caption,subcaption}
\usepackage{nicefrac}       % compact symbols for 1/2, etc.
\usepackage{microtype}      % microtypography
\usepackage[inline]{enumitem} %for enumerate in the same line
\usepackage{todonotes}
\usepackage[normalem]{ulem}

\usepackage{multirow}
\makeatletter
\usepackage{tikz}
\usetikzlibrary{shapes.misc, positioning}
\usetikzlibrary{backgrounds,fit,matrix, calc}
\tikzset{
   box/.style = {minimum height=10pt, minimum width=10pt, draw, rounded corners,rectangle, fill=white!50},
}
\usetikzlibrary{fadings}

\tikzset{
   boxconv/.style = {minimum height=2cm, minimum width=2cm, draw, line width=0.4mm, fill opacity=0.9, rounded corners,rectangle, fill=white!50},
}

\tikzset{
   boxconv_inactive/.style = {minimum height=2cm, minimum width=2cm,line width=0.3mm, draw, line width=0.1mm , fill opacity=0.9, rounded corners,rectangle, gray, fill=white!50},
}
\tikzset{
   input/.style = {minimum height=3cm, minimum width=3cm, draw, , fill opacity=0.9, rectangle, fill=white!50},
}

\tikzset{
   boxpooled/.style = {minimum height=1.5cm, minimum width=1.5cm, draw, line width=0.4mm, fill opacity=0.9, rounded corners,rectangle, fill=white!50},
}

\tikzset{
   boxpooled_inactive/.style = {minimum height=1.5cm, minimum width=1.5cm, draw, line width=0.4mm, fill opacity=0.9, rounded corners,rectangle, gray, fill=white!50},
}

\usetikzlibrary{fadings}\tikzset{
    boxwta/.style={%
        draw=black, thick,
        rectangle,
        rounded corners,
        minimum height=3cm,
        minimum width=3cm
    }
}

\usetikzlibrary{fadings}\tikzset{
    box1/.style={%
        draw=black, thick,
        rectangle,
        minimum height=2cm,
        minimum width=2cm
    }
}

\usetikzlibrary{fadings}\tikzset{
    box2/.style={%
        draw=black, thick,
        rectangle,
        minimum height=1.cm,
        minimum width=1.cm
    }
}

\usetikzlibrary{fadings}\tikzset{
    box3/.style={%
        draw=black, thick,
        rectangle,
        minimum height=.8cm,
        minimum width=.8cm
    }
}
\usetikzlibrary{decorations.markings}

\title{Local Competition and Uncertainty for Adversarial Robustness in Deep Learning}

\author{%
  Antonios Alexos\\
  Department of Electrical and Computer Engineering\\
  University of Thessaly\\
  Volos, Greece \\
  \texttt{aalexos@uth.gr} \\
   \And
   Konstantinos P. Panousis\\
   Dept. of Informatics and Telecommunications\\
   National and Kapodistrian University of Athens\\
   Athens, Greece \\
   \texttt{kpanousis@di.uoa.gr} \\
  \And
   Sotirios Chatzis\\
   Dept. of Electrical Eng., Computer Eng., and Informatics\\
   Cyprus University of Technology\\
   Limassol, Cyprus \\
   \texttt{sotirios.chatzis@cut.ac.cy} \\
}

\begin{document}

\maketitle

\begin{abstract}

This work attempts to address adversarial robustness of deep networks by means of novel learning arguments. Specifically, inspired from results in neuroscience, we propose a local competition principle as a means of adversarially-robust deep learning. We argue that novel local winner-takes-all (LWTA) nonlinearities, combined with posterior sampling schemes, can greatly improve the adversarial robustness of traditional deep networks against difficult adversarial attack schemes. We combine these LWTA arguments with tools from the field of Bayesian non-parametrics, specifically the stick-breaking construction of the Indian Buffet Process, to flexibly account for the inherent uncertainty in data-driven modeling. As we experimentally show, the new proposed model achieves high robustness to adversarial perturbations on MNIST and CIFAR10 datasets. Our model achieves state-of-the-art results in powerful white-box attacks, while at the same time retaining its benign accuracy to a high degree. Equally importantly, our approach achieves this result while requiring far less trainable model parameters than the existing state-of-the-art.

\end{abstract}

\section{Introduction}

In recent years, Deep Neural Networks (DNNs) have provided a variety of breakthroughs in numerous applications, especially in the Computer Vision Community, e.g., \cite{he2016deep, szegedy2013deep, badrinarayanan2017segnet}. However, it is a well-known fact, that DNNs are highly susceptible to adversarial attacks. In the adversarial context, \textit{adversarial examples}, i.e., inputs comprising carefully design perturbations, aim to ``\textit{fool}'' the considered model into \textit{misclassification}. Even small perturbations in the original input, e.g. an $L_p$ norm, can successfully render the model vulnerable, highlighting the frailness of the commonly employed DNN approaches in more demanding tasks \cite{papernot2017practical}. It is apparent, that this vulnerability, restricts their safe and confident employment in safety-critical applications, such as, autonomous driving \cite{boloor2019simple, chen2015deepdriving, kurakin2016adversarial}, video recognition \cite{jiang2019black}, healthcare \cite{finlayson2019adversarial} and other real-world scenarios \cite{kurakin2016adversarial}. To this end, significant research effort has been devoted in the Deep Learning community, in order to defend against various kinds of attacks, aspiring to make DNNs more robust to adversarial examples. Several approaches have been proposed to successfully tackle this task; however, each comes with its own disadvantages.

Adversarial attacks, as well as defense strategies, comprise many different approaches, sharing the same goal; making deep architectures more \textit{reliable} and \textit{robust}. In general, the adversarial defenses can be categorized as: (i) \textit{Adversarial Training}, where a model is trained both with the original, as well as perturbed data, aspiring to make the model more robust during inference \cite{madry2017towards,tramer2017ensemble,shrivastava2017learning}, (ii) \textit{Manifold Projections}, where the original data are projected into a different subspace presuming that therein, the effects of the perturbations can be mitigated \cite{jalal2017robust, shen2017ape, song2017pixeldefend} (iii) \textit{Stochasticity}, where some randomization of the input data and/or of the neuronal activations of each hidden layer is performed \cite{prakash2018deflecting, dhillon2018stochastic, xie2017mitigating}, and (iv) \textit{Preprocessing}, where some aspect of either the data or of the neuronal activations are modified or transformed \cite{buckman2018thermometer, guo2017countering, kabilan2018vectordefense}. 

%Some of the notable approaches belonging to the aforementioned categories, include: (1) the \textit{Defensive Distillation} (\cite{papernot2016distillation}), where one can either transfer or use extracted knowledge from different models \cite{carlini2017towards}, aiming to improve the model's robustness to various adversarial attacks, and (2) \textit{Feature Squeezing} (\cite{xu2017feature}), where the inputs features are reduced in order to compare the difference between the predictions of the original input and the reduced one after squeezing \cite{he2017adversarial}.

%Even though these approaches have provided some encouraging results, they do not come with any guarantees of generality and overall reliability, restricting their successful employment them in safety-critical tasks. In some cases, the provided guarantees (if any) may be excessively restricting, e.g. only for very small $L_p$ magnitudes. Not only that, usually, the proposed defense approaches are particularly tailored to a specific kind of adversarial attack. Thus, more often than not, when the adversarial attack patterns change, the considered models fail completely. 

% Moreover, many of the currently considered approaches, despite being able to defend against some kinds of attacks, may fail to successfully defend against others, e.g., being able to defend against white-box or black-box attacks but not both.

However, many of the currently considered approaches and architectures are particularly tailored to tackle a specific attack, restricting their generality to other potential adversarial schemes; thus, it is quite often the case that when the adversarial attack patterns change, the considered models fail completely. In this context, even though the commonly employed non-linearities, such as the Rectified Linear Units (ReLUs), provide a flexible computational tool for efficient training of DNNs, they do not exhibit any useful properties that may address the adversarial scenario; to overcome this deficiency of common neuronal activations, we may need to consider a radically different approach. 

Recently, there is a new invigorated interest in DL community in the creation of more biologically plausible models. Indeed, there is an increasing body of evidence in the neuroscience community, that neurons in a biological system with similar functions, are aggregated together in groups and local competition takes place for their activations. Thus, in each block only one neuron can be active at a given time, while the rest are inhibited to silence \cite{kandel2000principles, andersen1969participation, eccles2013cerebellum, stefanis1969interneuronal, douglas2004neuronal, lansner2009associative}, leading to a Local-Winner-Takes-All (LWTA) mechanism. Employing this mechanism to neural networks has been shown to provide promising results, introducing the ability to discover effective sparsely distributed representation of their inputs \cite{lee1999learning, olshausen1996emergence}, while exhibiting automatic gain control, noise suppression and robustness to catastrophic forgetting \cite{srivastava2013compete, grossberg1982contour, carpenter1988art}. These inherent properties of the competition-based mechanism, render LWTAs a potentially powerful biologically-inspired approach in the adversarial framework.

On the other hand, the inevitable overfitting tendencies of DNNs render them brittle against adversarial attacks; even small perturbations of the input data may easily fool an overfitted model. To this end, significant research effort has been devoted in attacking overfitting, aiming to develop methods in order to account for the \textit{uncertainty} of the considered architectures. Thus, many regularization methods have been proposed in the literature, e.g., Dropout \cite{gal2016dropout}. In this context, Chatzis \cite{chatzis2018indian} presented a novel approach, relying on the sparsity-inducing nonparametric Indian Buffet Process Prior (IBP) \cite{ghahramani2006infinite} in order to regularize the resulting architecture in a data-driven fashion. Using a set of auxiliary Bernoulli random variables, the utility of each architectural component was explicitly modeled, in an on-off fashion, intelligently inferring the necessary network complexity to represent the data. These approaches act as a source of inspiration for effectively addressing the adversarial scenario. We posit that the robustness of the model against adversarial examples may significantly increase, by accounting for the modeling uncertainty, while at the same time exploiting the benefits of the induced regularization.

Drawing upon these insights, we propose a new deep network design scheme that is particularly tailored to address the adversarial context. This capacity is effectuated via the combination of the LWTA mechanism and the nonparametric Bayesian IBP prior, allowing for potently addressing uncertainty. Moreover, we combine our proposed paradigm with Error Correcting Output Codes \cite{verma2019error}. 

We evaluate our approach using well-known benchmark datasets and architectures. The provided empirical evidence vouch for the potency of our approach, yielding state-of-the-art robustness against powerful white-box-attacks.

The remainder of the paper is organized as follows: In Section 2, we introduce the necessary theoretical background. In Section 3, we introduce the proposed approach and describe its integrated IBP and competition-based mechanisms. In Section 4 we perform extensive experimental evaluations, providing insights for the behavior of the produced model, while in Section 5, we summarize the contribution of this work, and discuss potential future directions.

\section{Theoretical Background}
In the following, we briefly present both the two core components utilized in this work, namely the nonparametric Indian Buffet Process Prior and the Local Winner-Takes-All mechanism.

\subsection{Indian Buffet Process}
The Indian Buffet Process (IBP) \cite{ghahramani2006infinite} defines a probability distribution over infinite binary matrices. IBP can be used as a flexible prior, allowing the number of considered features to be unbounded and inferred in a data-driven fashion. Its construction, ensures sparsity in the obtained representation, while at the same time allowing for more features to emerge, as new observations appear. Here, we focus on the stick-breaking construction of the IBP proposed by \cite{teh2007stick}, making the IBP amenable to Variational Inference. Let us consider $N$ observations, and a binary matrix $\boldsymbol Z = [z_{i,k}]_{i,k=1}^{N,K}$; each entry therein, indicates the existence of feature $k$ in observation $i$. Taking the infinite limit $K\rightarrow \infty$, we can construct the following hierarchical representation \cite{teh2007stick}:
\begin{align}
\label{ibp_equations}
u_k \sim \mathrm{Beta}(\alpha,1) \qquad \pi_k = \prod_{i=1}^k u_i \qquad z_{i,k} \sim \mathrm{Bernoulli}(\pi_k) \forall i 
\end{align}
where $\alpha$ is non-negative parameter, called the strength or innovation parameter, and controls the induced sparsity.
In practice, $K$ is set to be equal to the input dimensionality to avoid the overcomplete feature representation when $K \rightarrow \infty$.

%%%%%%%%%%%%%%%%%%%%%%%%%%%%%%%%%%%%%%%%%%%%%%%%
%%%%%%%%%%%% LOCAL WINNER TAKE ALL %%%%%%%%%%%%%
%%%%%%%%%%%%%%%%%%%%%%%%%%%%%%%%%%%%%%%%%%%%%%%%
\subsection{Local Winner-Takes-All}

Even though the commonly employed activations, such as ReLUs, constitute a convenient mathematical tool for training deep neural networks, they do not come with biological plausibility. In the following, we describe the general architecture when locally competing units are employed via the Local Winner-Takes-All (LWTA) mechanism. 

Let us assume a single layer of an LWTA-based network comprising $K$ LWTA blocks with $U$ competing units therein. Each block produces an output $\boldsymbol{y}_k$, $k=1, \dots, K$ given some input $\boldsymbol x \in \mathbb{R}^{N\times J}$. Each linear unit in each block, computes its activation $h_k^u, \ u=1, \dots U$, and the output of the each block is decided via competition. Thus, for each block $k$ and unit $u$ therein, the output reads:
\begin{align}
y_k^u = g(h_k^1, \dots, h_k^U)
\end{align}
where $g(\cdot)$ is the \textit{competition function}. The activation of each individual neuron follows the conventional inner product computation $h_k^u = w_{ku}^T \boldsymbol{x}$, where $\boldsymbol{W} \in \mathbb{R}^{J \times K \times U}$ is the weight matrix in this context. In the rigid LWTA network definition, the final output reads:
\begin{align}
y_k^u = 
\begin{cases}
1, & \text{if } h_k^u \geq h_k^i, \qquad \forall i=1, \dots, U, \ i \neq u\\
0, & \text{otherwise}
\end{cases}
\end{align}
To bypass the restrictive binary output, more expressive versions of the competition function have been proposed in the literature, e.g., \cite{srivastava2013compete}:
\begin{align}
y_k^u = 
\begin{cases}
h_k^u, & \text{if } h_k^u \geq h_k^i, \qquad \forall i=1, \dots, U, \ i \neq u\\
0, & \text{otherwise}
\end{cases}
\end{align}
It is apparent that, only the neuron with the \textit{strongest} activation produces an output in each block, while the others are inhibited to silence, i.e., the zero value. In this way, the output of each layer of the network yields a sparse representation according to the competition outcome within each block. Both these competition functions are called the \textit{hard} winner-takes-all function; in case of multiple winners the tie can be broken either by index or randomly. Panousis et al. \cite{panousis2019nonparametric}, proposed a novel competition function, based on a competitive random sampling procedure, driven by the activations of each neuron in each block. In our work, we adopt the latter, explained in detail in the following.

%\begin{figure}
%	\centering
%	\resizebox{0.7\linewidth}{!}{\input{figures/wta}}
%	\caption{A graphical representation of a competition-based architecture. Rectangles denote LWTA blocks and circles the competing units therein. The winner units are denoted with bold contours ($\xi=1$). Bold edges in the last layer, denote retained connections ($z=1$).}
%	\label{fig:feedforward}
%\end{figure}

%%%%%%%%%%%%%%%%%%%%%%%%%%%%%%%%%%%%%%%%%%%%%
%%%%%%%%%%%%%%%% PROPOSED MODEL %%%%%%%%%%%%%
%%%%%%%%%%%%%%%%%%%%%%%%%%%%%%%%%%%%%%%%%%%%%

\section{Model Definition}
\label{proposed_model_section}

In this work, we consider a principled way of designing deep neural networks that are particularly tailored to handle adversarial attacks. To this end, we combine the LWTA approach with appropriate arguments from the nonparametric Bayesian statistics. We aim to provide a more robust approach towards adversarial examples by flexibly accounting for the modeling uncertainty.

Let us assume an input dataset $ \boldsymbol{X} \in \mathbb{R}^{N\times J}$ with $N$ examples, comprising $J$ features each. In conventional deep architectures, each hidden layer comprises nonlinear units; the input is presented to the layer, which then computes an affine transformation via the inner product of the input with weights $\boldsymbol{W} \in \mathbb{R}^{J \times K}$, producing outputs $ \boldsymbol{Y} \in \mathbb{R}^{N \times K}$. The described computation for each example $n$ yields $\boldsymbol y_n = \sigma(\boldsymbol{x}_n W + b) \in \mathbb{R}^K, \ n=1, \dots, N$, where $b\in \mathbb{R}^K$ is a bias factor and $\sigma(\cdot)$ is a non-linear activation function, e.g. ReLU. An architecture comprises intermediate and output layers. We begin with the definition of an intermediate layer of our approach.

In the LWTA setting, the aforementioned procedure is modified; singular units are replaced by LWTA blocks, each containing a set of competing units. Thus, the layer input is now presented to each different block and each unit therein, via different weights. Assuming $K$ number of LWTA blocks and $U$ number of competing units, the weights are now represented via a three-dimensional matrix $\boldsymbol W \in \mathbb{R}^{J \times K \times U}$.

As previously mentioned, we follow \cite{panousis2019nonparametric}, where the local competition in each block is performed via a competitive random sampling procedure. Specifically, we define additional discrete latent vectors $\boldsymbol \xi_n \in \mathrm{one\_hot}(U)^K$, in order to encode the outcome of the local competition between the units in each block. For each datapoint $n$, the non-zero entry of the one-hot representation, denotes the winning unit among the $U$ competitors in each of the $K$ blocks of the layer.

To further account for the uncertainty and regularization of the resulting model, we turn to the nonparametric Bayesian framework. Specifically, we introduce a matrix of latent variables $Z \in \{0, 1\}^{J \times K }$, to explicitly regularize the model by inferring the utility of each synaptic weight connection in each layer. Each entry therein is set to one, if the $j^{th}$ dimension of the input is presented to the $k^{th}$ block, otherwise $z_{j,k} = 0$. We impose the sparsity-inducing IBP prior over the latent variables Z and perform inference over them.

We can now define the output of a layer of the considered model, $\boldsymbol{y}_n \in \mathbb{R}^{K\cdot U}$, as follows:
\begin{align}
[\boldsymbol{y}_n]_{ku} = [\boldsymbol \xi_n ]_{ku} \sum_{j=1}^J (w_{j,k,u} \cdot z_{j,k}) \cdot [ \boldsymbol{x}_n]_j \in \mathbb{R}
\label{eqn:layer_output_ff}
\end{align}
In order to facilitate a competitive random sampling procedure in a data-driven fashion, the latent indicators $\boldsymbol \xi_n$ are drawn from a posterior Categorical distribution that reads:
\begin{align}
q([\boldsymbol \xi_n]_k ) = \mathrm{Discrete} \left([\boldsymbol \xi_n]_k \Big | \mathrm{softmax}\left( \sum_{j=1}^J [w_{j,k,u}]_{u=1}^U \cdot z_{j,k} \cdot [\boldsymbol x_n]_j \right) \right)
\end{align}
The posteriors of the latent variables $Z$ are drawn from Bernoulli posteriors, such that:
\begin{align}
q(z_{j,k}) = \mathrm{Bernoulli}(z_{j,k} | \tilde{\pi}_{j,k})
\end{align}
Consequently, we impose a symmetric Discrete prior over the latent indicators, $[\boldsymbol \xi_n]_k \sim \mathrm{Discrete}(1/U)$, while we resort to fixed-point estimation for the weight matrices $\boldsymbol W$. The definition of an intermediate layer of the considered approach is now complete. 

For the output layer of our approach, we consider a similar variant based on the conventional feedforward layer. That is, we perform the standard inner product computation, while utilizing the IBP to further account for model uncertainty. 

Specifically, we assume an input $\boldsymbol x \in \mathbb{R}^{N \times J}$ to a $C$-unit output layer with weights $W \in \mathbb{R}^{J \times C}$. We introduce an analogous auxiliary matrix of latent variables $Z \in \{ 0, 1\}^{J \times C}$. Thus, the computation for the output $\boldsymbol{y}\in \mathbb{R}^{N \times C} $ yields:
\begin{align}
y_{n,c}=\sum_{j=1}^{J}\left(w_{j, c} \cdot z_{j, c}\right) \cdot\left[\boldsymbol{x}_{n}\right]_{j} \in \mathbb{R}
\end{align}
where the posterior latent variable $Z$ are drawn independently from a Bernoulli distribution:
\begin{equation}\label{Bernouli_posteriors}
q\left(z_{j, c}\right)=\operatorname{Bernoulli}\left(z_{j, c} | \tilde{\pi}_{j, c}\right)
\end{equation}
The prior for $Z$, once again follows the SBP of the IBP, while we seek fix-point estimates for the weights. The formulation of the full network architecture is now complete. A graphical illustration of the proposed approach is depicted in Fig. \ref{synopsis:wta}.

%%%%%%%%%%%%%%%%%%%%%%%%%%%%%%%%%%%%
%%%%%%%%%%%% Convolutional %%%%%%%%%
%%%%%%%%%%%%%%%%%%%%%%%%%%%%%%%%%%%
\subsection{Convolutional Layers}

In order to accommodate architectures comprising convolutional operations, we adopt the LWTA convolutional variant as defined in \cite{panousis2019nonparametric}. Specifically, let us assume an input tensor $\{ \boldsymbol{X}\}_{n=1}^N \in \mathbb{R}^{H \times L \times C}$ at a specific layer, where $H, L, C$ are the height, length and channels of the input. We define a set of kernels, each with weights $\boldsymbol{W}_k \in \mathbb{R}^{h \times l \times C \times U}$, where $h,l,C, U$ are the kernel height, length, number of channels and competing features maps, and $k=1, \dots K$. Thus, analogously to the grouping of linear units in the dense layers, in this case, local competition is performed among feature maps. Each kernel is treated as an LWTA block and each layer comprises multiple kernels competing for their outputs. We additionally consider an analogous auxiliary binary matrix $Z\in \{0,1\}^K$ to further regularize the convolutional layers.

\begin{figure}
	\begin{subfigure}[t]{.45\textwidth}
		\centering
		\resizebox{1.2\linewidth}{!}{\def\layersep{3.5cm}
\def\inputsize{2}
\def\wtablocks{2}
\def\neuronsep{5}
\def\outputsize{2}
\def\wtasep{2.5}
\def\wtablocks{3}
\def\unitsperblock{2}
\def\prob{0.55}

\begin{tikzpicture}[-,draw, node distance=\layersep, label/.style args={#1#2}{%
    postaction={ decorate,
    decoration={ markings, mark=at position #1 with \node #2;}}}]
    \tikzstyle{every pin edge}=[<-,shorten <=1pt]
    \tikzstyle{neuron}=[circle,draw,minimum size=12pt,inner sep=0pt]
    \tikzstyle{wta} = [rounded corners,rectangle, draw, minimum_height=3cm, minimum_width=2cm]

    \tikzstyle{input neuron}=[neuron, fill=white!50];
    \tikzstyle{output neuron}=[neuron, fill=white];
    \tikzstyle{hidden neuron}=[neuron, fill=white!50];
    \tikzstyle{hidden neuron activated}=[neuron, fill=white!50, very thick];
    \tikzstyle{wta block} =[wta];
    \tikzstyle{annot} = [text width=10em, text centered]

    % Draw the input layer nodes
    \node[input neuron, pin=left:$x_1$] (I-1) at (0,-\neuronsep+\wtasep) {};
    \node[input neuron, pin=left:$x_J$] (I-2) at (0,-2*\neuronsep+\wtasep) {};
		
    % Draw the hidden layer nodes
     	\matrix (H-1) at (\layersep, -\wtasep-0.*\wtasep) [row sep=4mm, column sep=2mm, inner sep=3mm, box, matrix of nodes] 
      {
        		\node[hidden neuron activated](o1-1){}; \\
        		\node[hidden neuron](o1-2){}; \\
	  };

	  %\matrix (H-2) at (\layersep, -2*\wtasep-0.*\wtasep) [row sep=4mm, column sep=2mm, inner sep=2mm, box, matrix of nodes] 
      %{
      %  		\node[hidden neuron](o2-1){}; \\
      %  		\node[hidden neuron activated](o2-2){}; \\
	  %};
	  
	  \matrix (H-2) at (\layersep, -3*\wtasep-0.*\wtasep) [row sep=4mm, column sep=2mm, inner sep=3mm, box, matrix of nodes] 
      {
        		\node[hidden neuron activated](o2-1){}; \\
        		\node[hidden neuron](o2-2){}; \\
	  };

	% Draw the second layer
	\matrix (H2-1) at (2*\layersep, -\wtasep-0.*\wtasep) [row sep=4mm, column sep=2mm, inner sep=3mm, box, matrix of nodes] 
      {
        		\node[hidden neuron](o21-1){}; \\
        		\node[hidden neuron activated](o21-2){}; \\
	  };

	  \matrix (H2-2) at (2*\layersep, -3*\wtasep-0.*\wtasep) [row sep=4mm, column sep=2mm, inner sep=3mm, box, matrix of nodes] 
      {
        		\node[hidden neuron](o22-1){}; \\
        		\node[hidden neuron activated](o22-2){}; \\
	  };
	  
%	  \matrix (H2-3) at (2*\layersep, -3*\wtasep-0.*\wtasep) [row sep=4mm, column sep=2mm, inner sep=2mm, box, matrix of nodes] 
%      {
%        		\node[hidden neuron activated](o23-1){}; \\
%        		\node[hidden neuron](o23-2){}; \\
%	  };
        
    %%%%%%%%%%%%%%%%%%%%%%%%%%%%
    %%%%%%%%% OUTPUT %%%%%%%%%%%
    %%%%%%%%%%%%%%%%%%%%%%%%%%%%
    % Draw the output layer node
    \foreach \name / \y in {1,...,\outputsize}
    		\node[output neuron] (O-\name) at (3*\layersep,-\neuronsep*\y+\wtasep) {};

    %%%%%%%%%%%%%%%%%%%%%%%%%%%%
    %%%%%%%%% PATH %%%%%%%%%%%%%
    %%%%%%%%%%%%%%%%%%%%%%%%%%%%
    \path (I-1) -- (I-2) node [black, font=\Huge, midway, sloped] {$\dots$};
 	\path[black!100, line width=0.55pt, thick] (I-1.east) edge node[ above, rotate=5, black!100] {\scriptsize $z_{1,1}=1$} (o1-1);
    \path[black!100, line width=0.55pt, thick] (I-1.east) edge node[ below, rotate=-15, black!100] {\scriptsize $z_{1,1}=1$} (o1-2.west);
    
    \path[black!100, line width=0.55pt, thick] (I-1.east) edge (o2-1.west);
    \path[black!100, line width=0.55pt, thick] (I-1.east) edge (o2-2.west);

	\path[black!100, line width=0.55pt, thick] (I-2.east) edge (o1-1.west);
    \path[black!100, line width=0.55pt, thick] (I-2.east) edge (o1-2.west);
    
    \path[black!40] (I-2.east) edge node[above, rotate=5, black!100] {\scriptsize $z_{J,K}=0$} (o2-1.west);
    \path[black!40] (I-2.east) edge node[below, rotate=-15, black!100] {\scriptsize $z_{J,K}=0$} (o2-2.west);
    
    	%\path[black!70, semithick] (I-3) edge (o1-1);
    %\path[black!70, semithick] (I-3) edge (o1-2);
    
    %\path[black!20, semithick] (I-3) edge (o2-1);
    %\path[black!20, semithick] (I-3) edge (o2-2);
            
    \path (H-1) -- (H-2) node [black, font=\Huge, midway, sloped] {$\dots$};
    \path (H2-1) -- (H2-2) node [black, font=\Huge, midway, sloped] {$\dots$};

    \path[color=black!100, line width=0.55pt, thick] (o1-1.east) edge (o21-1.west); 
    %node[ above] {\scriptsize $z_{1,1}=1$}   
    \path[color=black!100, line width=0.55pt, thick] (o1-1.east) edge (o21-2.west);

    \path[color=black!40] (o1-1.east) edge (o22-1.west);  
    \path[color=black!40] (o1-1.east) edge (o22-2.west); 
    
    \path[black!40] (o1-2.east) edge (o21-1.west);  
    \path[black!40] (o1-2.east) edge (o21-2.west); %node[ below, color=black!100] {\scriptsize %$z_{2,1}=0$} 

    %\path[densely dotted] (o3-2) edge (o21-1);  
    %\path[densely dotted] (o3-2) edge (o21-2);
    
    \path[black!40] (o1-2.east) edge (o22-1.west);  
    \path[black!40] (o1-2.east) edge (o22-2.west);

    \path[color=black!100, line width=0.55pt, thick] (o2-1.east) edge (o21-1.west);  
    \path[color=black!100, line width=0.55pt, thick] (o2-1.east) edge (o21-2.west);
    
    \path[color=black!100, line width=0.55pt, thick] (o2-1.east) edge (o22-1.west);  
    \path[color=black!100, line width=0.55pt, thick] (o2-1.east) edge (o22-2.west);
    
    \path[black!40] (o2-2.east) edge (o21-1.west);  
    \path[black!40] (o2-2.east) edge (o21-2.west);
    
    \path[black!40] (o2-2.east) edge (o22-1.west);  
    \path[black!40] (o2-2.east) edge (o22-2.west);
    
    \path (O-1) -- (O-2) node [black, font=\Huge, midway, sloped] {$\dots$};

    %\path (o1-1) edge (o23-1);  
    %\path (o1-1) edge (o23-2); 
    
    %\path[densely dotted] (o2-1) edge (o23-1);  
    %\path[densely dotted] (o2-1) edge (o23-2);
    
    %\path (o2-1) edge (o23-1);  
    %\path (o2-1) edge (o23-2);
    
    % second units in blocks

    %\path[densely dotted] (o3-2) edge (o22-1);  
    %\path[densely dotted] (o3-2) edge (o22-2);
    
    %\path[densely dotted] (o1-2) edge (o23-1);  
    %\path[densely dotted] (o1-2) edge (o23-2); 
    
    %\path (o2-2) edge (o23-1);  
    %\path (o2-2) edge (o23-2);
    
    %\path[densely dotted] (o3-2) edge (o23-1);  
    %\path[densely dotted] (o3-2) edge (o23-2);

    %%%%%%%%%%%%%%%%%%%%
    % Second to output%%
    %%%%%%%%%%%%%%%%%%%%
    \path[black!40] (o21-1.east) edge (O-1.west);  
    \path[black!40] (o21-1.east) edge (O-2.west);
    %\path[black!40] (o21-1) edge (O-3);
    
    \path[black!100, line width=0.55pt, thick] (o21-2.east) edge (O-1.west);  
    \path[black!100, line width=0.55pt, thick] (o21-2.east) edge (O-2.west);

    \path[black!40] (o22-1.east) edge (O-1.west);  
    \path[black!40] (o22-1.east) edge (O-2.west);
    %\path[black!40] (o22-1) edge (O-3);
    
    \path[black!100, line width=0.55pt, thick] (o22-2.east) edge (O-1.west);  
    \path[black!100, line width=0.55pt, thick] (o22-2.east) edge (O-2.west);

    %\path[black!40] (o23-1) edge (O-1);  
    %\path[black!40] (o23-1) edge (O-2);
    %\path[black!40] (o23-1) edge (O-3);
    
    %\path[black!40] (o23-2) edge (O-1);  
    %\path[black!40] (o23-2) edge (O-2);
    %\path[black!40] (o23-2) edge (O-3);

    \node[annot,above= -1mm of o1-1] (k) {\scriptsize $\xi$= 1};
    \node[annot,below= -1mm of o1-2] (k) {\scriptsize $\xi$= 0};
    
    \node[annot,above= -1mm of o22-1] (k) {\scriptsize $\xi$= 0};
    \node[annot,below= -1mm of o22-2] (k) {\scriptsize $\xi$= 1};
    % Annotate the layers
    \node[annot,above of=H-1, node distance=1.8cm] (hl) {SB-LWTA layer};
    \node[annot,above left=-3mm and -1.7cm of H-1] (k) {\scriptsize $1$};
    
    \node[annot,above left=-3mm and -1.7cm of H2-1] (k) {\scriptsize $1$};
    
    \node[annot,below left=-3mm and -1.7cm of H-2] (k) {\scriptsize $K$};
    
    \node[annot,below left=-3mm and -1.7cm of H2-2] (k) {\scriptsize $K$};
    
    \node[annot,above of=H2-1, node distance=1.8cm] (hl2) {SB-LWTA layer};
    \node[annot,left of=hl] {Input layer};
    \node[annot,right of=hl2] {Output layer};
\end{tikzpicture}
% End of code
}
		\caption{}
		\label{synopsis:wta}
	\end{subfigure}
	\begin{subfigure}[t]{.55\textwidth}
		\centering
		\resizebox{.95\linewidth}{!}{\input{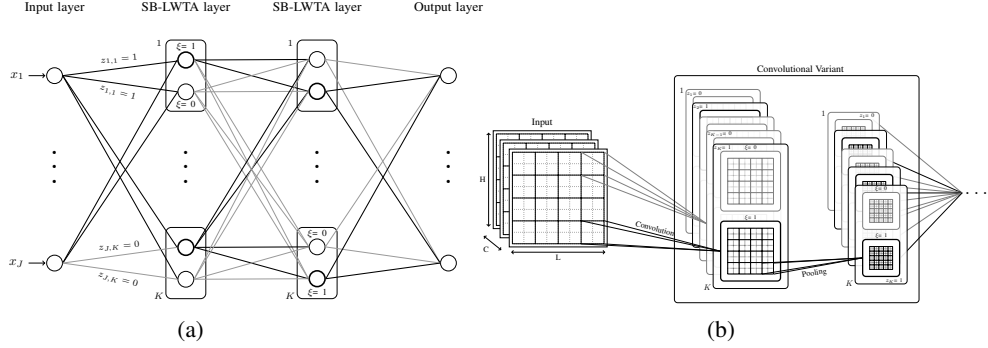}}
		\caption{}
		\label{synopsis:fig:cnn_sb_lwta}
	\end{subfigure}
	\caption{ (a) A graphical representation of a competition-based architecture. Rectangles denote LWTA blocks and circles the competing units therein. The winner units are denoted with bold contours ($\xi=1$). Bold edges in the last layer, denote retained connections ($z=1$). (b) The convolutional LWTA variant. Competition takes places among feature maps. The winner feature map (denoted with bold contour) passes its output to the next layer, while the rest are zeroed out. }
\end{figure}

Thus, at a given layer of the corresponding convolutional variant, the output $\boldsymbol Y_n \in \mathbb{R}^{H \times L \times K \cdot U}$ is obtained via concatenation along the last dimension of the subtensors:
\begin{align}
[\boldsymbol Y_n]_k = [ \boldsymbol \xi_n]_k \left( \boldsymbol (z_k \cdot W_k) \star \boldsymbol X_n \right) \in H \times L \times U
\end{align}
where $\boldsymbol X_n$ is the input tensor for the $n^{th}$ datapoint, ``$\star$'' denotes the convolution operation and $[\boldsymbol \xi_n]_k \in \mathrm{one\_hot}(U)$ is a one-hot vector representation with $U$ components.  Turning to the competition function, we follow the same rationale, such that the sampling procedure is driven from the outputs of the competing feature maps:
\begin{align}
q([\boldsymbol{\xi}_n]_k) = \mathrm{Discrete}([\boldsymbol \xi_n]_k \Big | \mathrm{softmax}(\sum_{h',l'} [ \boldsymbol (z_k \cdot W_k) \star \boldsymbol X_n]_{h',l',u})
\end{align}
We impose an IBP prior on $Z$, while the posteriors are drawn from a Bernoulli distribution, such that,  $q(z_k) = \mathrm{Bernoulli}(z_k | \tilde{\pi}_k)$. We impose an analogous symmetric prior for the latent winner indicators $ [\boldsymbol{\xi}_n]_k \sim \mathrm{Discrete}(1/U)$. A graphical illustration of the defined layer is depicted in Fig. \ref{synopsis:fig:cnn_sb_lwta}.

%\begin{figure}
%	\centering
%	\resizebox{0.75\linewidth}{!}{\input{figures/wta_conv.tex}}
%	\caption{The convolutional LWTA variant. Competition takes places among feature maps. The winner feature map (denoted with bold contour) passes its output to the next layer, while the rest are zeroed out. The layer can be followed by conventional operations, such as pooling.}
%	\label{fig:convolutional}
%\end{figure}
	
%%%%%%%%%%%%%%%%%%%%%%%%%%%%%%%%%%%%%%%%
%%%%%%%% COMPONENT UTILITY %%%%%%%%%%%%%
%%%%%%%%%%%%%%%%%%%%%%%%%%%%%%%%%%%%%%%%
\subsection{Training \& Inference}
To train the proposed model, we resort to maximization of the Evidence Lower Bound (ELBO). To facilitate efficiency in the resulting procedures, we adopt the Stochastic Gradient Variational Bayes framework (SGVB) \cite{kingma2014autoencoding}. However, our model comprises latent variables that are not readily amenable to the reparameterization trick of SGVB, namely, the discrete latent variables $Z$ and $\boldsymbol \xi$, and the Beta-distributed stick variables $\boldsymbol u$. To this end, we utilize the continuous relaxation based on the Gumbel-Softmax trick \cite{maddison2016concrete,jang} for the discrete random variables and the Kumarawamy distribution \cite{kumaraswamy} as an approximation to the Beta distribution. These approximations are only employed during training. At inference, we draw samples from the respective original distribution of the latent variables, i.e. the Discrete, Bernoulli and Beta distributions. We employ the mean-field assumption to facilitate efficient optimization.

Differently than similar approaches, the uncertainty modeling of our approach is based on two different sampling processes. On the one hand, contrary to the rigid competition function in the context of competition-based networks presented in \cite{srivastava2013compete}, we implement a data-driven random sampling procedure to determine the winning units. On the other, we further account for the uncertainty of the learning process by sampling from a Bernoulli latent variable based on an IBP prior. Through the introduced latent variables, we can regularize the model by exploiting the sparsity inducing behavior of the IBP prior and the overall induced regularization via the maximization of the resulting ELBO. 

%%%%%%%%%%%%%%%%%%%%%%%%%%%%%%%%%%
%%%%%%%% EXPERIMENTAL %%%%%%%%%%%%
%%%%%%%%%%%%%%%%%%%%%%%%%%%%%%%%%%

\section{Experimental Evaluation}

We evaluate the capacity of our proposed approach against various adversarial attacks and under different setups. We follow the experimental framework as in \cite{verma2019error}. To this end, we employ a modified decoding error-correcting output code strategy to the output of the proposed network, as described therein, and we examine the resulting modeling capabilities of an LWTA and IBP based DNN, both in classification accuracy, as well as in the classification confidence levels.

\subsection{Implementation Details}

Let us denote by $\boldsymbol{C}$ an $ M \times N$ \textit{coding matrix}, where $M$ is the number of classes and $N$ is the codeword length. Therein, the $k^{th}$ row, denotes the desired output of the network, when the input is from class $k$; in this work, we consider only cases with $N=M$ or $N>M$. We employ coding matrices of varying length and explore multiple ways of transforming the logits of the last layer of the network to probabilities, including the Softmax, the Logistic, and the Hyperbolic Tangent Function. Both standard and ensemble architectures are considered \cite{verma2019error}. We resort to SGVB for training the model, utilizing the Kumaraswamy (\cite{kumaraswamy})  and continuous relaxation of the Discrete distribution proposed in \cite{jang, maddison2016concrete}.

\subsection{Experimental Setup}

We consider two popular benchmark datasets for adversarial research, namely, MNIST \cite{lecun2010mnist} and  CIFAR10 \cite{Krizhevsky09learningmultiple}. We employ the \textit{same architectures}, as in \cite{verma2019error}, employing both Standard and Ensemble architectures. We consider two different splits for our approach: (i) an architecture comprising LWTA blocks with 2 competing units, and (ii) one with 4 competing units. In the case of ensemble architectures containing layers with less than 4 overall units, we employ a hybrid of our approach using both 2 and 4 way splits. We examine 5 different kinds of adversarial attacks: (i) Projected Gradient Descent (PGD), (ii) Carlini and Wagner (CW), (iii) Blind Spot Attack (BSA) (\cite{zhang2019limitations}), (iv) a random attack (Rand) \cite{verma2019error}, and (v) an attack comprising additive uniform noise corruption ($+U(-1,1)$). See the Supplementary Material for the detailed implementation details and experimental setup.

\subsection{Experimental Results}

In the following, we present the comparative experimental evaluations of our approach. In the provided tables, we select a subset of four different models (as defined in \cite{verma2019error}) employing our proposed approach, which we compare with the Madry model \cite{madry2017towards}\, and the best-performing  TanhEns16 and TanhEns64 models of \cite{verma2019error}, in order to unclutter the presentation of the experimental results. Detailed presentation of all the considered models and methods is included in the Supplementary.

\textbf{MNIST}. For the MNIST dataset, we train the network for a maximum of 50 epochs, using the same data augmentation as in \cite{verma2019error}. In Table \ref{MNIST_results}, the comparative results for each different architecture and adversarial attack are depicted. As we observe, our approach yields considerable improvements over the Madry \cite{madry2017towards} and TanhEns16 \cite{verma2019error} architectures in three of the five considered attacks, while employing architectures with lower computational footprint. The differences in the rest of the experiments are negligible. For example, consider the Softmax model comprising LWTA blocks with 2 competing units and TanhEns16 model, which requires $\approx 20\%$ more parameters. In the benign classification accuracy, we observe a $0.004\%$ difference in favor of TanhEns16, while the Softmax LWTA model exhibits an improvement of $0.055\%$ in the PGD and a significant $15.9\%$ improvement in the uniform corruption noise attack.  The empirical evidence suggest that our approach yields better performance than the alternatives, even when employing models with less parameters.

\begin{table}[ht!]
 \centering
 \resizebox{0.85\linewidth}{!}{
 \begin{tabular}{c c c c c c c c}
 \hline
 \addlinespace[0.2cm]
   Model & Params & Benign & PGD & CW & BSA & Rand & $+\mathrm{U}(-1,1)$\\ 
%   Model & Parameters & {\parbox{1.3cm}{\centering Benign \\ Accuracy}} & {\parbox{1.3cm}{\centering PGD \\ ($\epsilon=0.3$)}} & CW & {\parbox{1.35cm}{\centering BSA \\($\alpha=0.8$)}} & {\parbox{1.3cm}{\centering Rand \\ (Pr.<0.9)}} & $+\mathrm{U}(-1,1)$\\ 
  \addlinespace[0.2cm]
 \hline
 \addlinespace[0.1cm]
 Softmax (U=4) & 327,380 & 0.9613 & 0.935 & 0.97 &  0.95 &  0.187 & 0.929\\
 Softmax (U=2) & 327,380 & 0.9908 & \textbf{0.984} & 0.97 & \textbf{1.0} & 0.961 & \textbf{0.986}\\
 Logistic (U=2) & 327,380 & 0.9749 & 0.963 & 0.96 & 0.96 & 0.998 & 0.963\\
%  Tanh16(U=2) & ??? & 0.9891 & 0.981 & 0.96 & 0.98 & 0.853 & 0.988\\
 LogEns10 (U=2) & \textbf{205,190} & 0.9653 & 0.946 & 0.95 & 0.94 & \textbf{1.0}& 0.949\\
 Madry \cite{madry2017towards} & 3,274,634 & 0.9853 & 0.925 & 0.84 & 0.52 & 0.351 & 0.15\\
 TanhEns16 \cite{verma2019error} & 401,168 & \textbf{0.9948} & 0.929 & \textbf{1.0} & \textbf{1.0} & 0.988 & 0.827\\

\addlinespace[0.1cm]
\hline
\addlinespace[0.2cm]
\end{tabular}
}
\centering
\caption{Accuracy scores for various models and adversarial attacks on the MNIST dataset. All considered architectures for our approach are the the same as the ones proposed in \cite{verma2019error}. Here, we report the best-performing adaptations. }
\label{MNIST_results}
\end{table}

%%%%%%%%%%%%%%%%%%%%%%%%%%%%%%%%%%%%%%%%%
%%%%%%%%%%%%%%%% CIFAR 10 %%%%%%%%%%%%%%
%%%%%%%%%%%%%%%%%%%%%%%%%%%%%%%%%%%%%%%%

\textbf{CIFAR-10.} For the CIFAR-10 dataset, we follow an analogous procedure, utilizing the same architectures and procedures as in \cite{verma2019error}. In Table \ref{CIFAR_results}, the obtained comparative effectiveness of our approach is depicted. In this case, the differences in the computational burden are more evident, since CIFAR-10 requires more involved architectures, considering its inherent complexity both for benign classification, as well as, for adversarial attacks. As in the previous experiments, we observe that our method presents significant improvements in three of the considered adversarial attacks, namely PGD, CW and BSA, utilizing networks with \textit{significantly less parameters} (even $2$ to $4$ times) than the best performing alternative of \cite{verma2019error}.

\begin{table}[ht!]
 \centering
  \resizebox{0.85\linewidth}{!}{
 \begin{tabular}{c c c c c c c c}
 \hline
 \addlinespace[0.2cm]
  Model & Params & Benign & PGD & CW & BSA & Rand & $+\mathrm{U}(-1,1)$\\ 
  \addlinespace[0.2cm]
 \hline
 \addlinespace[0.1cm]
 Tanh16(U=4) & 773,600 & 0.5097 & 0.46 & 0.55 & 0.6 & 0.368 & 0.436\\
 Softmax(U=2) & \textbf{772,628} & 0.8488 & 0.83 & \textbf{0.85} & \textbf{0.83} & 0.302 & 0.825\\
 Tanh16(U=2) & 773,600 & 0.8539 & 0.826 & 0.83 & \textbf{0.83} & 0.32 & 0.815\\
 LogEns10(U=2) & 1,197,998 & 0.8456 & \textbf{0.846} & 0.83 & 0.8 &  \textbf{1.0} & 0.812\\
 Madry \cite{madry2017towards} & 45,901,914 & 0.871 & 0.47 & 0.08 & 0 & 0.981 & 0.856\\
 TanhEns64 \cite{verma2019error} & 3,259,456 & \textbf{0.896} & 0.601 & 0.76 & 0.76 & \textbf{1.0} & \textbf{0.875}\\
\addlinespace[0.1cm]
\hline
\addlinespace[0.2cm]
\end{tabular}
}
\centering
\caption{Accuracy scores for various models and adversarial attacks on the CIFAR10 dataset. Once again, we consider all the architectures proposed in \cite{verma2019error} and report the best-performing adaptations. }
\label{CIFAR_results}
\end{table}

\subsection{Further Insights}

\begin{figure}[!tbp]
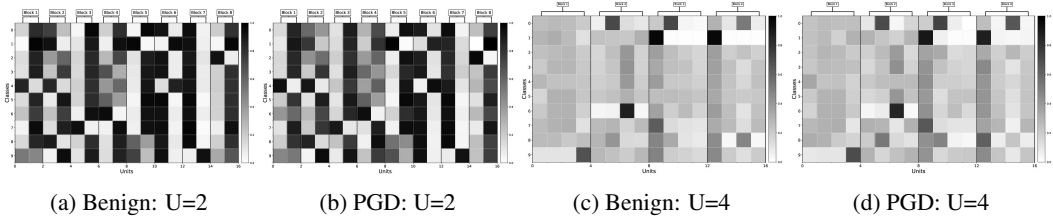

  \centering
  \begin{subfigure}[b]{0.24\textwidth}
  	\includegraphics[width=\textwidth]{figures/unit_probs/cifar_benign_tanh_U_2.pdf}
  	\caption{Benign: U=2}
  	\label{probs_benign_U_2}
  \end{subfigure}
%   \hfill
  \begin{subfigure}[b]{0.24\textwidth}
    \includegraphics[width=\textwidth]{figures/unit_probs/cifar_pgd_tanh_U_2.pdf}
    \caption{PGD: U=2}
    \label{probs_pgd_U_2}
  \end{subfigure}
%   \hfill
  \begin{subfigure}[b]{0.25\textwidth}
    \includegraphics[width=\textwidth]{figures/unit_probs/cifar_benign_tanh16_U_4.pdf}
    \caption{Benign: U=4}
    \label{probs_benign_U_4}
  \end{subfigure}
%   \hfill
  \begin{subfigure}[b]{0.25\textwidth}
    \includegraphics[width=\textwidth]{figures/unit_probs/cifar_pgd_tanh16_U_4.pdf}
    \caption{PGD: U=4}
    \label{probs_pgd_U_4}
  \end{subfigure}
 \caption{Probabilities of competing units in LWTA blocks, of an intermediate layer of the Tanh16 model, for each class in the CIFAR-10 dataset. Fig. \ref{probs_benign_U_2} and Fig. \ref{probs_pgd_U_2}, depict the activations of a model layer with 2 competing units, for benign and PGD test examples respectively. Figs. \ref{probs_benign_U_4} and \ref{probs_pgd_U_4} correspond to a network layer comprising 4 competing units. Black denotes very high winning probability, while white denotes very low probability.}
\label{probabilities_figure}
\end{figure}

We scrutinize the behavior of the LWTA mechanism in the performed experimental evaluations, in order to gain some insights by examining the resulting competition patterns, and assess that the competition does not collapse to singular ``always-winning'' units. To this end, we choose a random intermediate layer of a Tanh16 model comprising 8 or 4 LWTA blocks of 2 and 4 competing units respectively, and focus on the CIFAR-10 dataset. The probabilities of unit activations for each class are depicted in Fig. \ref{probabilities_figure}. For the former, the probabilities for benign test examples are illustrated in Fig. \ref{probs_benign_U_2}, while the corresponding probabilities for the PGD method, are presented in Fig.\ref{probs_pgd_U_2}. As we observe, the unit activation probabilities for each different setup are essentially the same, suggesting that the LWTA mechanism, succeeds in encoding salient discriminative patterns in the data, while also exhibiting in practice, the inherent property of noise suppression. The empirical evidence suggest that we can obtain networks with strong generalization value, able to defend against adversarial attacks in a principled way. In Fig. \ref{probs_benign_U_4} and \ref{probs_pgd_U_4}, the corresponding probabilities when employing 4 competing units are depicted. We observe that in this case, the competition is uncertain about the winning unit in each block and for each class, exhibiting an average activation probability for each unit around $\approx 50\%$. Moreover, there are several differences in the activations between the benign data and a PGD attack, explaining the significant drop in performance. This behavior potentially arises due to the relatively small structure of the network; from the 16 units of the original architecture, only 4 are active for each input. Thus, in this setting, LWTA fails to encode the necessary distinctive patterns of the data.  Analogous illustrations for different setups are provided in the Supplementary.

Table \ref{time_results} depicts the inference times for The Softmax model, on various attacks on MNIST datasets. We compare the Softmax model with our proposed architecture, to the architecture in \cite{verma2019error}. We observe that our proposal, is \textbf{4.3 times faster} for the execution of PGD attack. For the CW attack it is \textbf{4.35 times faster}, while for the Random and the Uniform Noise attack it is \textbf{3 times faster}.

\begin{table}[ht!]
 \centering
  \resizebox{0.85\linewidth}{!}{
 \begin{tabular}{c c c c c c c c}
 \hline
 \addlinespace[0.2cm]
  Model & Benign & PGD & CW & BSA & Rand & $+\mathrm{U}(-1,1)$\\ 
  \addlinespace[0.2cm]
 \hline
 \addlinespace[0.1cm]
 Softmax  & 7.969 & 108.219 & 767.292 & 823.25 & 1.370 & 13.534\\
 Softmax\cite{verma2019error} & 2.218 & 48.294 & 3322.743 & 3581.75 & 3.703 & 40.89\\
\addlinespace[0.1cm]
\hline
\addlinespace[0.2cm]
\end{tabular}
}
\centering
\caption{Inference times for various attacks for the Softmax model with 2 competing units on the MNIST dataset.}
\label{time_results}
\end{table}

\begin{figure}[ht!]
	\centering % <-- added
	\medskip
	\begin{subfigure}{0.33\textwidth}
		\centering
		\includegraphics[width=0.9\linewidth]{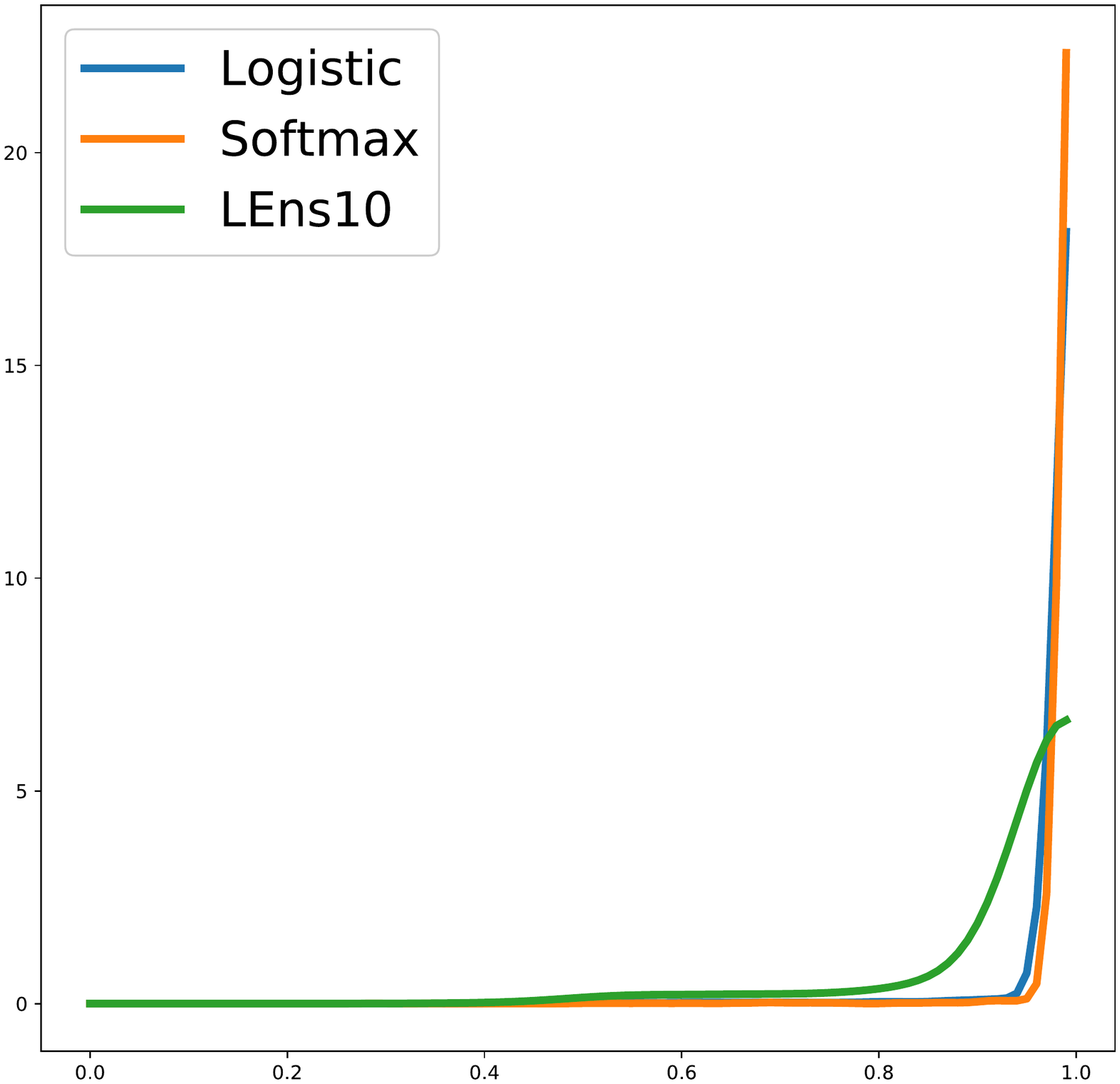}
		\caption{Benign}
		\label{mnist_benign_U_2}
	\end{subfigure}\hfil % <-- added
	\begin{subfigure}{0.33\textwidth}
		\centering
		\includegraphics[width=0.9\linewidth]{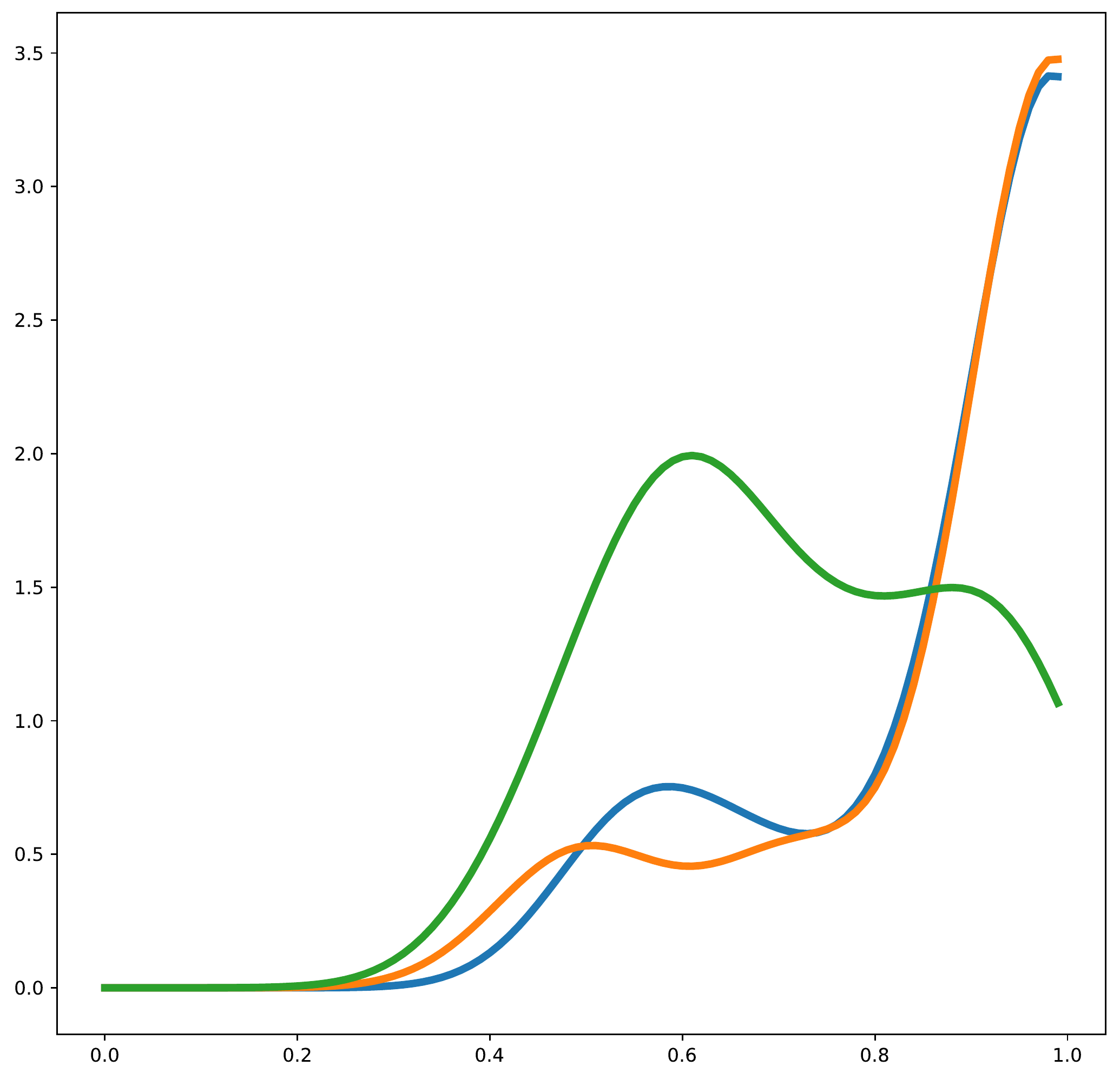}
		\caption{PGD}
		\label{mnist_pgd_U_2}
	\end{subfigure}\hfil % <-- added
	\begin{subfigure}{0.33\textwidth}
		\centering
		\includegraphics[width=0.9\linewidth]{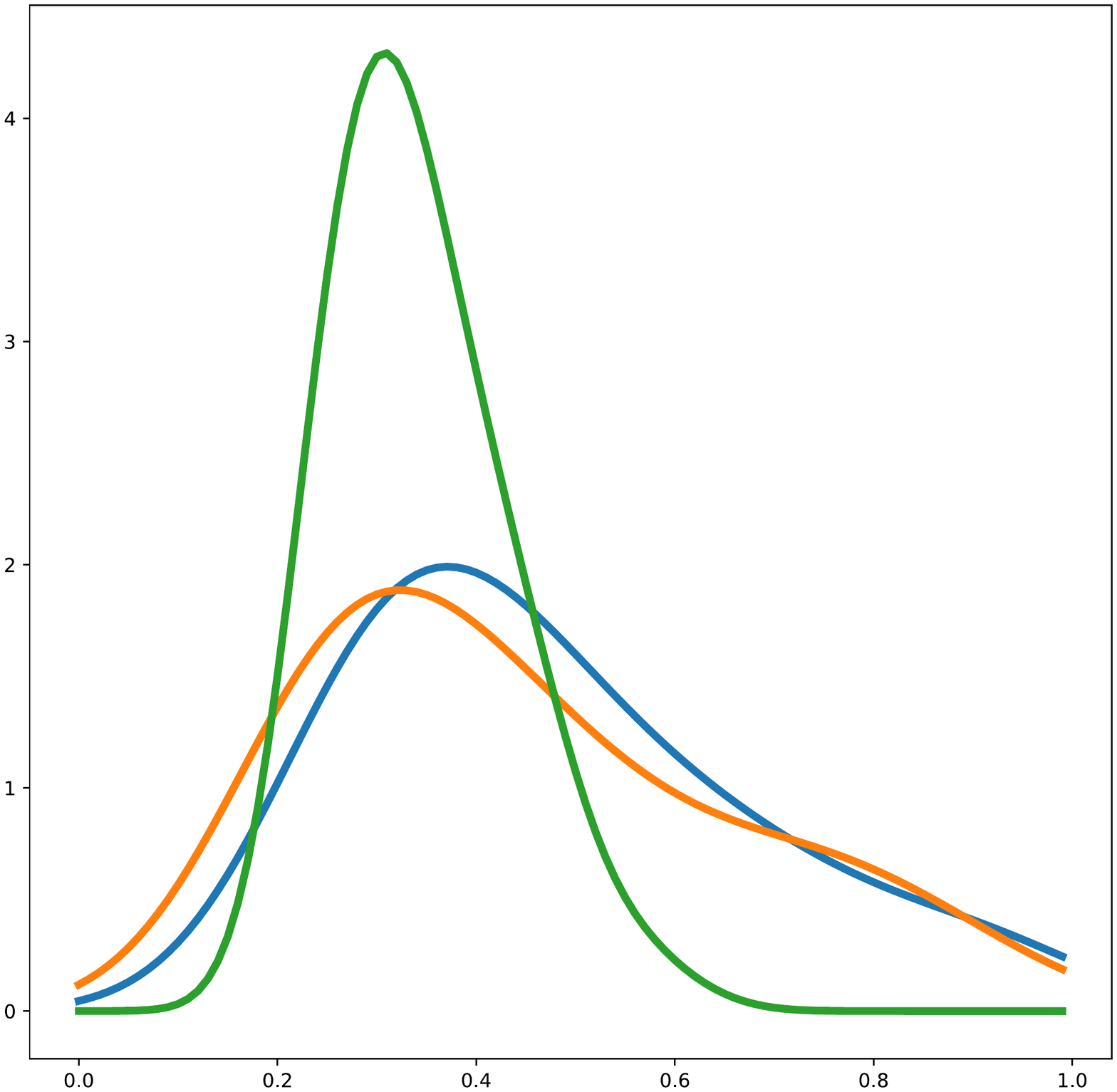}
		\caption{Random}
		\label{mnist_random_U_2}
	\end{subfigure}
	\caption{Probability distributions assigned to the most probable class on the test-set of MNIST by various models. LEns10 refers to the LogisticEns10 model.}
	\label{distributions_U_2}
\end{figure}

Finally, we examine the uncertainty estimation capabilities of the introduced approach via the resulting confidence in the classification task. To this end, and similar to \cite{verma2019error}, we compare the probability distributions of class assignment over a randomly chosen set of test examples of the MNIST dataset for our best performing models with $U=2$. Since our approach yields high classification accuracy in the benign case, we expect that the considered models should exhibit high confidence in label assignment. Indeed, we observe the expected behavior in Fig. \ref{mnist_benign_U_2}. Fig. \ref{mnist_pgd_U_2},  depicts the models behavior on a PGD attack. We observe that in some cases the models are (correctly) less confident about the class label due to the adversarial perturbations. However, and contrary to \cite{verma2019error, madry2017towards}, since our approach retains high classification accuracy, the models assign substantial probability mass in high confidence levels. The empirical evidence suggests, that by employing a careful blend of LWTA activations and the IBP, we can overcome the ``irrational''  overconfidence of the softmax activation. Softmax \textit{incorrectly} assigns high confidence to the class with the largest logits, even in the presence of adversarial examples. Thus, we can overcome this flaw, even when employing small length error correcting output codes. Fig. \ref{mnist_random_U_2} illustrates the resulting behavior in randomly generated inputs. In this case, all models correctly place most of the probability mass in the low spectrum. The experimental results vouch for the efficacy of our approach, flexibly accounting the inherent uncertainty in the adversarial context, thus retaining high classification rate against various attacks. The corresponding graphs for the CIFAR-10 dataset can be found in the Supplementary.

% \tonycomment{We give emphasis to the sparsity of SBP on the output, by using a small length of error correcting output codes, which proves its efficiency and superiority to aforementioned codes. Moreover, LWTA and SBP help the model to overcome the irrational overconfidence of the softmax activation. Softmax assigns higher probability to the class whose corresponding logit is the largest, which means that it is certain almost everywhere in logit space. Thus, it is also confident on adversarial inputs, for which the model is incorrect. As you can see from the results our proposed Softmax Model, we overcome this flaw.}

\section{Conclusions and Future Work}
In this paper, we proposed a novel approach towards adversarial robustness in deep networks, aiming to create networks that exhibit more robust and reliable behavior to white-box adversarial attacks. To this end, we utilized local competition between linear units, in the form of the Local-Winner-Takes-All mechanism, combined with the sparsity inducing IBP prior in order to account for the uncertainty and regularize the learning process. Our experimental evaluations have provided strong empirical evidence for the efficacy of our approach. The proposed method exhibits considerable improvements in various kinds of adversarial attacks, while retaining high accuracy in the benign context. It is noteworthy, that the competition mechanism yielded very similar competition patterns for benign and adversarial examples, suggesting that the considered approach succeeds in encoding differential essential sub-distributions of the data through specialization. In our future work, we aim to explore the potency of the LWTA and IBP integration in other challenging adversarial scenarios.

\bibliographystyle{neurips_2020}

\end{document}